\documentclass[journal]{IEEEtran}
\ifCLASSINFOpdf
\else
\fi

\usepackage{caption}
\usepackage{comment}

\begin{document}
%
\title{VoD: Learning Volume of Differences for Video-Based Deepfake Detection}


\author{
Ying Xu, ~\IEEEmembership{Student Member,~IEEE},
Marius Pedersen, ~\IEEEmembership{Member,~IEEE}, 
Kiran Raja, ~\IEEEmembership{Senior Member,~IEEE} \\
}

\maketitle

\begin{abstract}
The rapid development of deep learning and generative AI technologies has profoundly transformed the digital contact landscape, creating realistic Deepfake that poses substantial challenges to public trust and digital media integrity. This paper introduces a novel Deepfake detention framework, Volume of Differences (VoD), designed to enhance detection accuracy by exploiting temporal and spatial inconsistencies between consecutive video frames. VoD employs a progressive learning approach that captures differences across multiple axes through the use of consecutive frame differences (CFD) and a network with stepwise expansions. We evaluate our approach with intra-dataset and cross-dataset testing scenarios on various well-known Deepfake datasets. Our findings demonstrate that VoD excels with the data it has been trained on and shows strong adaptability to novel, unseen data. Additionally, comprehensive ablation studies examine various configurations of segment length, sampling steps, and intervals, offering valuable insights for optimizing the framework. The code for our VoD framework is available at {\url{https://github.com/xuyingzhongguo/VoD}}.
\end{abstract}


%
\IEEEpeerreviewmaketitle

\section{Introduction}
\label{sec:intro}
The digital content landscape has been significantly altered by the rapid advancement of deep learning and generative AI,  blurring the lines between reality and altered content~\cite{Heikkila2024, WEF2023, Radcliffe2023}. The ability to create Deepfake, which can alter a person's identity or manipulate their facial expressions and movements by deep learning methods, demonstrates the dual nature of this innovation. 
While it offers extensive possibilities for creativity and expression, it also poses significant risks. 
The misuse of Deepfake can enable digital impersonation and fraud, undermining public trust and leading to social and legal challenges~\cite{Goudemond2024, Boyd2022, Villasenor2023, WEF2023}. 
Considering these implications, there is an imperative need to develop robust and accurate systems for detecting Deepfake in order to maintain the integrity of digital media and protect societal trust. 

There are two main types of Deepfake detection mechanisms: frame-level detection and video-level detection, each employing distinct methodologies and technologies to identify forgeries. 
Frame-level detection primarily utilizes convolutional neural networks (CNNs) to analyze spatial patterns within individual frames, harnessing the network's ability to discern authenticity from manipulated facial features. 
This method benefits from a rich array of research, focusing on traditional spatial features~\cite{DBLP:conf/iccv/RosslerCVRTN19, afchar2018mesonet, wang2020cnn, nguyen2019capsule} and more nuanced aspects like image statistics~\cite{frank2020leveraging, tian2023frequency, wang2023noise} and semantic information~\cite{agarwal2019protecting, huang2023implicit}, which help reveal subtle inconsistencies often missed by straightforward visual analysis.

In contrast, video-level detection introduces the additional complexity of temporal dynamics, extending the detection capabilities beyond static images to sequences of frames. 
This approach can be divided into two primary methodologies.
The first integrates spatial-temporal neural networks to analyze frame volumes over time, extracting patterns that only emerge across sequences, thus capturing spatial and temporal anomalies~\cite{zhang2021detecting, gu2021spatiotemporal}.
The second focuses explicitly on temporal inconsistencies, such as irregular lip movement or facial expressions over time~\cite{haliassos2021lips, zheng2021exploring}, leveraging specialized networks designed to trace and analyze the continuity and coherence of facial dynamics throughout a video~\cite{liu2023ti2net}.

Despite the effectiveness of current Deepfake detection mechanisms at both the frame and video levels, notable limitations could pave the way for further research and innovation in this field. 
A significant challenge in Deepfake detection is the generalizability of existing methods to adapt to new and evolving forgery techniques. 
One reason for this issue could be that many current video-level Deepfake detection methods employ limited techniques for inputting videos, typically by simply concatenating cropped faces together~\cite{zhang2021detecting, gu2021spatiotemporal, zhang2021detecting}.
This kind of approach tends to omit critical temporal information about facial movements and inconsistencies between frames. Further, most of the existing works employ the differences in frames directly at the spatial level, and we assert the differences exist along multiple axes, which include spatial and temporal directions.

\begin{figure*}[htp]
\centering
\includegraphics[width=0.95\textwidth]{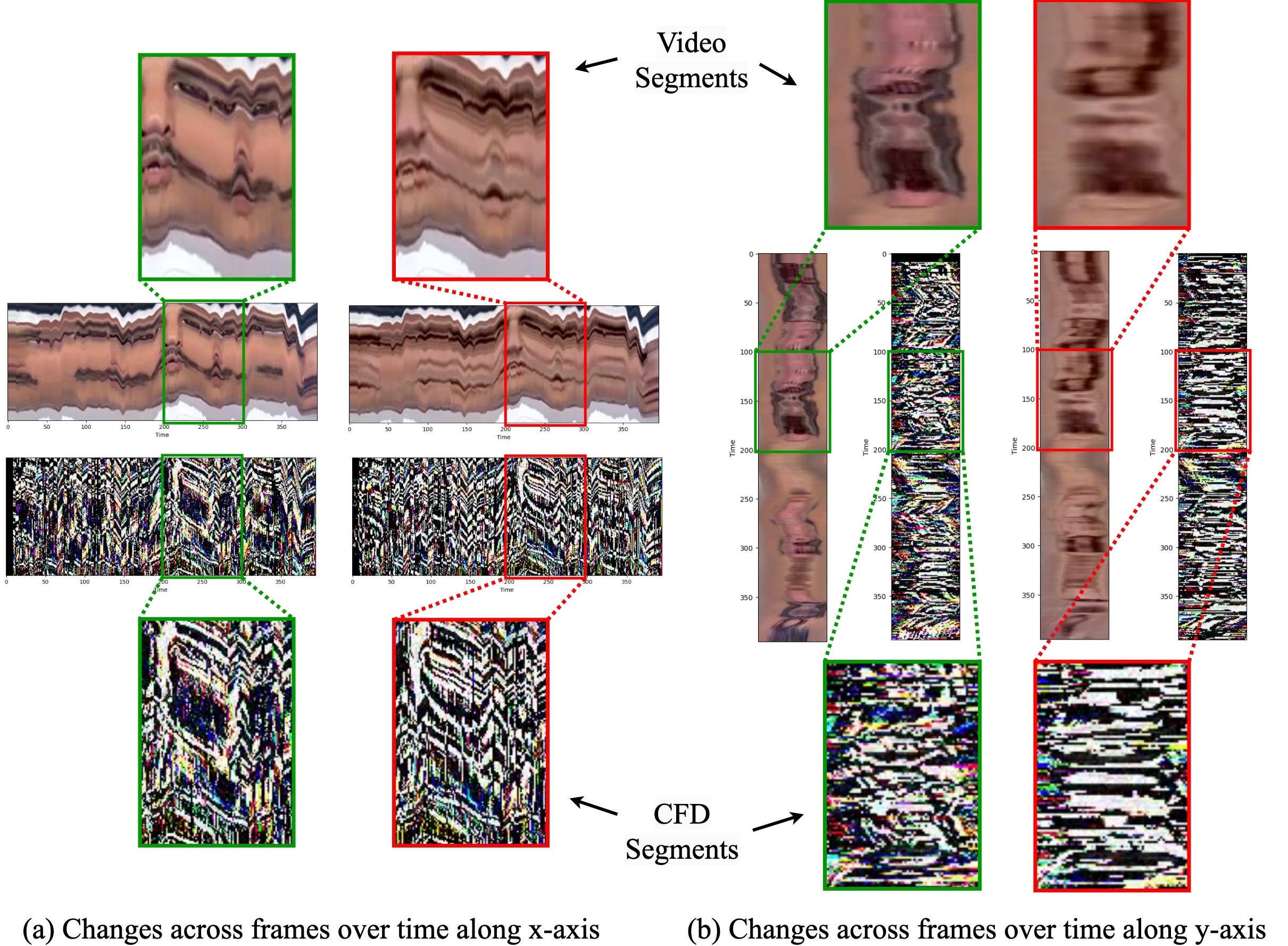}
\captionof{figure}{The changes over time for both vertical and horizontal lines through the video segments and the Consecutive Frame Differences~(CFD) segments of video. (a) Changes along the x-axis and (b) Changes along the y-axis. Each panel includes zoomed insets that compare real video sequences and CFD with their Deepfake counterparts. The insets with green borders are for real videos while the insets with red borders are for Deepfake videos.}
\label{fig:slice}
\end{figure*}

To address this gap, in this paper, we propose a new framework, Volume of Differences~(VoD), for Deepfake detection, which uses difference learning along spatial and temporal axes. We achieve this by first obtaining the Consecutive Frame Difference (CFD) which extracts temporal information between consecutive frames. Further, the obtained CFD segments are treated as a volume of differences to learn the subtle differences for classification. 
The differences in addition from frame to frame also exist in different axes such as $(x,y,t)$, which correspond to spatial $(x,y)$ and temporal $(t)$ axes as shown in \Cref{fig:slice}. 
To fully leverage the differences across these spatial and temporal axes between real and Deepfake videos, the differences are learnt using stepwise expansions that expand a single axis in each step ({\textit{i.e.}}, either in $x$, $y$ or $t$). Further, a simple supervised classifier is used with a fully connected layer for the final classification. The proposed video-based approach offers flexibility in configuring parameters such as segment length $C_{sl}$, sampling steps $C_{step}$, and sampling interval $C_{in}$. It also allows for various methods to calculate frame differences and the use of different backbone networks. By systematically varying these parameters, the proposed approach is evaluated for its effects on the performance of detecting Deepfake. Our findings demonstrate how optimized CFD configurations significantly enhance the performance, providing robust insights for more accurate and efficient Deepfake detection systems.

We evaluate our proposed VoD framework through a series of experiments conducted on the FaceForensics++ (FF++) dataset~\cite{DBLP:conf/iccv/RosslerCVRTN19}, which includes four distinct types of manipulated media: Deepfakes (DF)~\cite{ffdf}, Face2Face (F2F)~\cite{thies2016face2face}, FaceSwap (FS)~\cite{fffs}, and NeuralTextures (NT)~\cite{thies2019deferred}. 
Our findings indicate that our method achieves better results compared to state-of-the-art (SOTA) methods in intra-dataset scenarios.
Additionally, we explore the generalizability of our proposed method on unseen datasets, Celeb-DF (v2)~\cite{li2020celeb}, DFDC~\cite{dolhansky2020Deepfake}, and DeepFakeDetection (DFD)~\cite{google_research_dfd} in video-based detection settings. Furthermore, we conduct comprehensive ablation studies focused on the Consecutive Frame Difference (CFD) technique within our analysis framework.

\section{Related Work}
\label{sec:related-work}
Current Deepfake detection methods are primarily categorized into frame-level and video-level approaches~\cite{zheng2021exploring}. Frame-level detection scrutinizes individual frames for signs of manipulation~\cite{DBLP:conf/iccv/RosslerCVRTN19, afchar2018mesonet, wang2020cnn, nguyen2019capsule, frank2020leveraging, tian2023frequency, wang2023noise, agarwal2019protecting, huang2023implicit}, while video-level detection analyzes temporal information across a video sequence to identify forgeries~\cite{zhang2021detecting, gu2021spatiotemporal, haliassos2021lips, zheng2021exploring, liu2023ti2net}.

\subsection{Frame-level Deepfake Detection}
One prominent direction of the frame-level Deepfake detection method is the use of convolutional neural networks (CNNs) to capture spatial patterns in the generated images~\cite{DBLP:conf/iccv/RosslerCVRTN19, afchar2018mesonet, wang2020cnn, nguyen2019capsule}. 
These methods leverage the powerful feature extraction capabilities of CNNs to train binary classifiers that can differentiate between authentic and manipulated faces. 

\begin{figure*}[htb]
  \centering
  \includegraphics[width=0.85\textwidth]{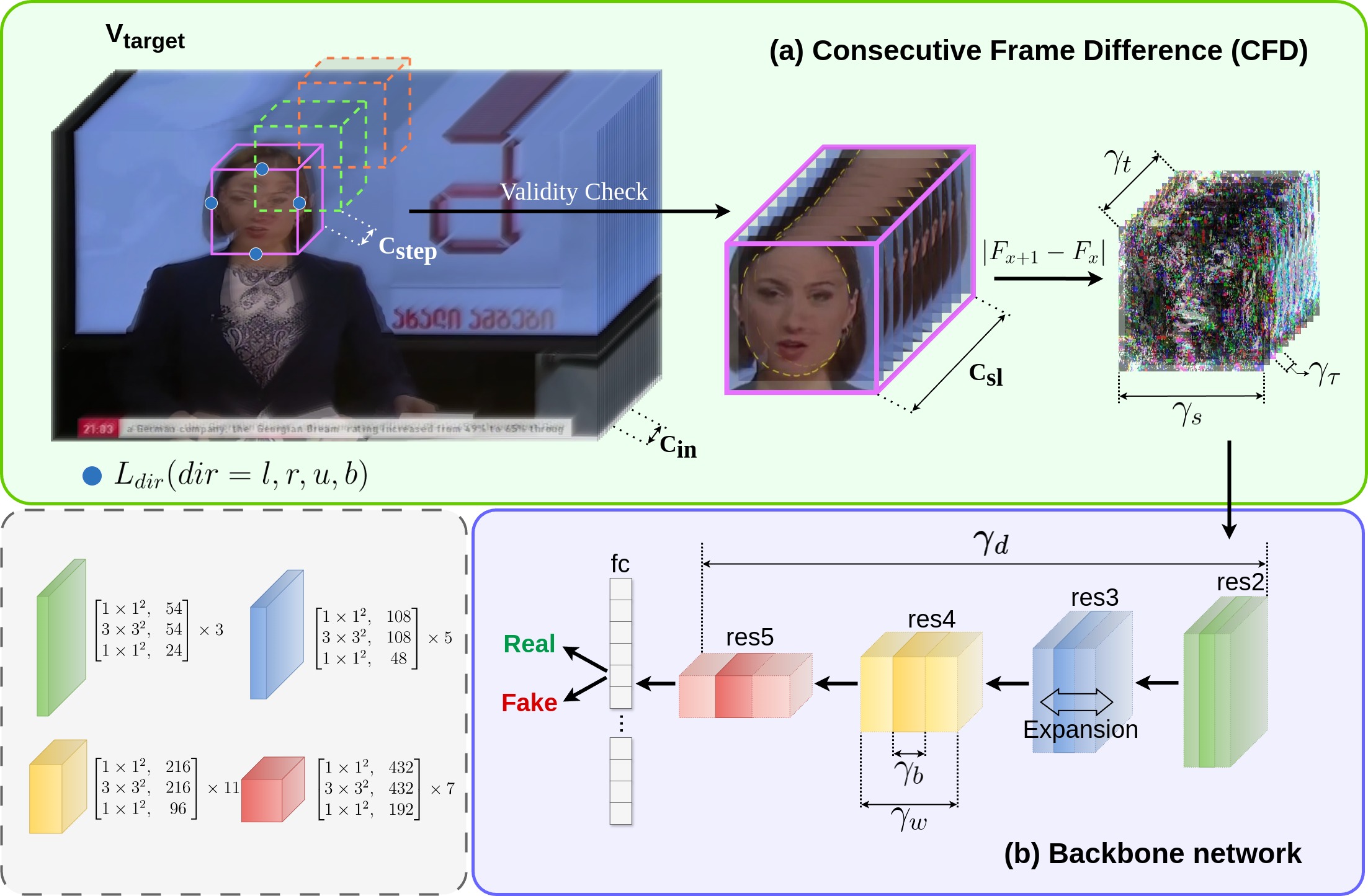}
  \caption{Schematic diagram of our proposed Volume of Differences~(VoD) framework. The target video $V_{target}$ is first provided to the Consecutive Frame Difference~(CFD) block to obtain the input segments with a given set of parameters such as segment length $C_{sl}$, sampling steps $C_{step}$, and sampling interval $C_{in}$. These segments are subsequently used to extract the features along each of the spatial and temporal axes using multiple sub-blocks $res$ and for the final classification. The detailed layers of the X3D backbone network used in the framework are shown in the dashed line box.}
  \label{fig:whole-network}
\end{figure*}

Beyond spatial analysis, a growing body of work has also explored the utilization of image statistics~\cite{frank2020leveraging, tian2023frequency, wang2023noise}, as well as semantic information~\cite{agarwal2019protecting, huang2023implicit}, for forgery detection. 
These approaches aim to uncover the subtle discrepancies between real and fake faces that may not be easily captured by spatial analysis alone.
For instance, Huang {\textit{et al.}}~\cite{huang2023implicit} considered a perspective for face-swapping detection that focuses on the implicit identity of faces using an identity-driven framework.
Wang {\textit{et al.}}~\cite{wang2023noise} presented a noise-based Deepfake detection model which focused on the underlying forensic noise traces left behind the Deepfake videos.

\subsection{Video-level Deepfake Detection}
Recently, many studies have started to consider the temporal dimension for face forgery detection at the video level~\cite{zhang2021detecting, gu2021spatiotemporal, haliassos2021lips, zheng2021exploring, liu2023ti2net}. 
There are two main directions in using temporal information.

One direction is to use spatial-temporal deep neural networks for forgery media classification.
Zhang {\textit{et al.}}~\cite{zhang2021detecting} fed the fixed-length frame volumes sampled from a video into a 3-dimensional Convolutional Neural Network to extract features across different scales and identify whether they are Deepfake. 
Gu {\textit{et al.}}~\cite{gu2021spatiotemporal} utilized a spatial inconsistency module and a temporal inconsistency module to capture spatial and temporal information individually before sending them to an information supplement module for classification.

Another direction is to focus on spotting the temporal inconsistency.
For instance, Haliassos {\textit{et al.}}~\cite{haliassos2021lips} introduced a method that focuses on identifying discrepancies in lip movements to detect face forgeries.
Zheng {\textit{et al.}}~\cite{zheng2021exploring} proposed a fully temporal convolution network along with a temporal transformer network to explore temporal coherence video forgery detection.
Liu {\textit{et al.}}~\cite{liu2023ti2net} proposed the Temporal Identity Inconsistency Network, a Deepfake detector that focuses on temporal identity inconsistency.

\subsection{Limitations of Current Works}
The generation of Deepfake media typically involves the application of sophisticated facial mapping and manipulation techniques on a frame-by-frame basis~\cite{DBLP:conf/iccv/RosslerCVRTN19, li2020celeb}. 
While post-processing methods such as anti-flickering, motion smoothing, and colour correction are often employed to enhance the visual coherence of the resulting footage~\cite{li2020celeb}, subtle temporal inconsistencies can still persist. 
These inconsistencies, as shown in \Cref{fig:slice}, can go undetected by models that rely solely on frame-level analysis~\cite{cao2022end}, as they are rooted in the underlying temporal dynamics of the video.

\section{Proposed Method}
\label{sec:method}

Our volume of differences~(VoD) framework, as shown in \Cref{fig:whole-network}, is designed to exploit the temporal and spatial incoherence between consecutive frames that can differ for videos that are real as compared to Deepfake. The temporal inconsistencies can emerge due to the Deepfake generation mechanism leading to abrupt frame transitions, incorrect alignment of faces or ghosting artifacts due to face manipulation. To effectively utilize such information, we first extract independent frames and crop the face area~\cite{DBLP:conf/iccv/BulatT17}. Noting the ghosting artifacts around the face silhouette and hair region in Deepfake videos, we allow the face cropping to extend the region of interest from the leftmost, rightmost, topmost, and bottom-most face landmarks, denoted as $L_{box}$, which includes $[L_l, L_r, L_u, L_b]$.
The faces in the segment are then cropped using $C_{sl}$, $F_s$, $C_{in}$, and $L_{box}$. Further, to utilize the differences across different frames, we subtract the frames that are consecutive using the CFD block discussed below. The obtained CFD segment is subsequently used to extract the features along each of the spatial and temporal axes using multiple sub-blocks indicated as $res$ in \Cref{fig:whole-network}. The features learnt along different axes are used to train a supervised classifier to distinguish between the two classes of real and Deepfake videos.

\subsection{Consecutive Frame Difference}
The consecutive frame differences (CFD) are used to obtain the frame level differences, which are further used to extract features. A target video $V_{target}$ is taken as an input to obtain the output CFD segments $S_{CFD}$. The output $S_{CFD}$, in essence, is a volume of frame differences governed by segment length $C_{sl}$; sampling steps $C_{step}$, and sampling interval $C_{in}$. The absolute differences between consecutive faces $|F_{x+1}-F_x|, x \in [1, C_{sl}-1]$ from the segment to complete the process of obtaining $S_{CFD}$ where $x$ is a given frame number in a particular video.


\subsection{Multi-axes Residual Learning for DeepFake Detection}
The CFD segment $S_{CFD}$ is used to learn the subtle residuals for distinguishing real videos from Deepfake ones. 
The $S_{CFD}$ contains subtle information along spatial and temporal axes, which, if leveraged, can help in detecting the manipulation in videos.
Considering the nature of the data, which is essentially a volume of differences, the proposed VoD approach employs an Expandable 3D Network (X3D) designed for video data processing~\cite{feichtenhofer2020x3d}.
The X3D model adopts a progressive network expansion strategy, initially starting from a tiny base 2D image architecture and methodically expanding into a spatio-temporal structure. 
This expansion occurs along multiple axes, including temporal duration $\gamma_t$, frame rate $\gamma_\tau$, spatial resolution $\gamma_s$, network width $\gamma_w$, bottleneck width $\gamma_b$ and depth $\gamma_d$, shown in \Cref{fig:whole-network}~(b).
While X3D offers multiple configurations, our proposed approach makes use of a small variant to save on computational complexity \cite{feichtenhofer2020x3d}, referred to as X3D-S, which consists of four different residual blocks which help in learning along multiple axes. The architecture configuration of X3D-S employs expansion of $\gamma_t=6$, $\gamma_\tau=13$, $\gamma_s=\sqrt{2}$, $\gamma_w=1$, $\gamma_b=2.25$ and $\gamma_d=2.2$. The residuals learnt along multiple axes are then used to learn a supervised classifier using a fully-connected layer with two classes ({\textit{i.e.}}, real and Deepfake).

\section{Experiments}
\label{sec:experiments}

\subsection{Experimental Setup}
To comprehensively assess the effectiveness of our proposed VoD framework, we conducted a series of experiments comparing it with current state-of-the-art techniques. 
This section provides details on the datasets used, the evaluation metrics applied, and the specific implementation details of our approach.

\textbf{Datasets.}
We evaluate our proposed VoD framework alongside current state-of-the-art techniques using the following datasets: FaceForensics++ (FF++)~\cite{DBLP:conf/iccv/RosslerCVRTN19}, Celeb-DF (v2)~\cite{li2020celeb}, DFDC~\cite{dolhansky2020Deepfake}, and DFD~\cite{google_research_dfd}. 
\textbf{FF++} dataset is a widely used benchmark that includes raw videos, high-quality (light compression version, constant rate quantization parameter equal to 23) and low-quality (c40) versions. 
It consists of 1000 real videos sourced from YouTube, along with corresponding fake videos created using four different manipulation techniques: Deepfakes (DF)~\cite{ffdf}, Face2Face (F2F)~\cite{thies2016face2face}, FaceSwap (FS)~\cite{fffs}, and NeuralTextures (NT)~\cite{thies2019deferred}. 
\textbf{Celeb-DF} dataset is a well-known benchmark for cross-dataset analysis, comprising 590 real videos and 5,693 DeepFake videos of celebrities. It features an enhanced compositing process to minimize various visual artifacts in the videos. 
\textbf{DFDC} is a large-scale benchmark created for the Deepfake Detection Challenge, containing 124,000 videos featuring 3,426 paid actors.
\textbf{DFD}, published by Google and Jigsaw, consists of Deepfake videos created from recordings of consenting actors. Its purpose is to support the development of synthetic video detection methods and aid the research community in mitigating the misuse of synthetic media.

\textbf{Implementation Details.} Following the official split of FF++, we used 740 videos for training, 140 for validation, and 140 for testing. 
For each training video, we extract facial segments with a length of 17 frames, with an interval of 4 frames, up to frame 200. Therefore, a maximum of 50 segments, each with a CFD segment with a length of 16, are used from each video. 
The size of the input is $224\times224$. 
The length of segments depends on the architectures of different 3D deep neural networks.
The entire network was trained using the Adam optimizer for 50 epochs, starting with a learning rate of $1e-2$ and a weight decay of $1e-4$. 
A step learning scheduler was employed to decrease the learning rate by a factor of 10 every 10 epochs.
All the experiments are conducted on NVIDIA P100 GPUs available.

\begin{table}[htb]
  \caption{FF++ Intra-dataset comparison. All the models are trained on FF++(c23) dataset.}
  \label{tab:ff-intra}
  \centering
  \renewcommand{\arraystretch}{1.3}
  \small
  \begin{tabular}{lcc}
    \hline
    \multirow{2}{*}{Methods} & \multicolumn{2}{c}{FF++(c23)} \\
    \cline{2-3}
                             & ACC & AUC \\
    \hline
    MFF~\cite{zhang2024face} & 94.14 & 98.78 \\
    RECCE~\cite{cao2022end} & 97.06 & 99.32 \\
    MultiAtt~\cite{zhao2021multi} & 97.60 & 99.29 \\
    M2TR~\cite{wang2022m2tr} & 97.93 & 99.51 \\
    VoD~(ours) & \textbf{98.77} & \textbf{99.53} \\
    \hline
  \end{tabular}
\end{table}

\subsection{Comparison with Existing Methods}
\label{subsec:result-compare}
We evaluate the effectiveness of the proposed VoD framework with both seen and unseen data in this section.

\subsubsection{Intra-dataset evaluation}
\Cref{tab:ff-intra} presents an intra-dataset comparison using FF++(c23) as the training data. 
The results demonstrate that our proposed VoD framework surpasses the state-of-the-art (SOTA) methods, achieving the highest accuracy of 98.77\% and an AUC score of 99.53.

\begin{table}[htb]
  \caption{Cross-testing in terms of AUC across four manipulation techniques used in the FF++ dataset. The grey background indicates within-dataset results.}
  \label{tab:ff1-3}
  \centering
  \renewcommand{\arraystretch}{1.3} 
  \begin{tabular}{lcccccc}
    \hline
    \textbf{Methods} & \textbf{Train} & \textbf{DF} & \textbf{F2F} & \textbf{FS} & \textbf{NT} & \textbf{Cross Avg.} \\
    \hline
    MFF~\cite{zhang2024face} & & \cellcolor{gray!20}98.74 & 62.17 & 64.10 & 61.84 & 71.71\\
    RECCE~\cite{cao2022end} & DF & \cellcolor{gray!20}\textbf{99.65} & \textbf{70.66} & 74.29 & \textbf{67.34} & \textbf{77.99} \\
    VoD~(ours) & & \cellcolor{gray!20}99.28 & 62.68 & \textbf{78.95} & 66.66 & 76.89\\
    \hline
    MFF~\cite{zhang2024face} & & 67.30 & \cellcolor{gray!20}95.41 & 58.45 & 59.36 & 70.13\\
    RECCE~\cite{cao2022end} & F2F & 75.99 & \cellcolor{gray!20}\textbf{98.06} & 64.53 & 72.32 & 77.73 \\
    VoD~(ours) & & \textbf{83.46} & \cellcolor{gray!20}98.04 & \textbf{74.95} & \textbf{73.88} & \textbf{82.58}\\
    \hline
    MFF~\cite{zhang2024face} & & 77.73 & 56.55 & \cellcolor{gray!20}98.15 & 53.20 & 71.41\\
    RECCE~\cite{cao2022end} & FS & 82.39 & 64.44 & \cellcolor{gray!20}98.82 & \textbf{56.70} & 75.59 \\
    VoD~(ours) & & \textbf{90.98} & \textbf{67.00} & \cellcolor{gray!20}\textbf{99.70} & 48.62 & \textbf{76.58}\\
    \hline
    MFF~\cite{zhang2024face} & & 74.91 & 69.71 & 53.75 & \cellcolor{gray!20}87.23 & 71.40 \\
    RECCE~\cite{cao2022end} & NT & 78.83 & 80.89 & \textbf{63.70} & \cellcolor{gray!20}93.63 & 79.26 \\
    VoD~(ours) & & \textbf{87.01} & \textbf{82.94} & 53.75 & \cellcolor{gray!20}\textbf{97.98} & \textbf{80.42}\\
    \hline
  \end{tabular}
\end{table}

\begin{figure}[htb]
  \centering
  \includegraphics[width=0.5\textwidth]{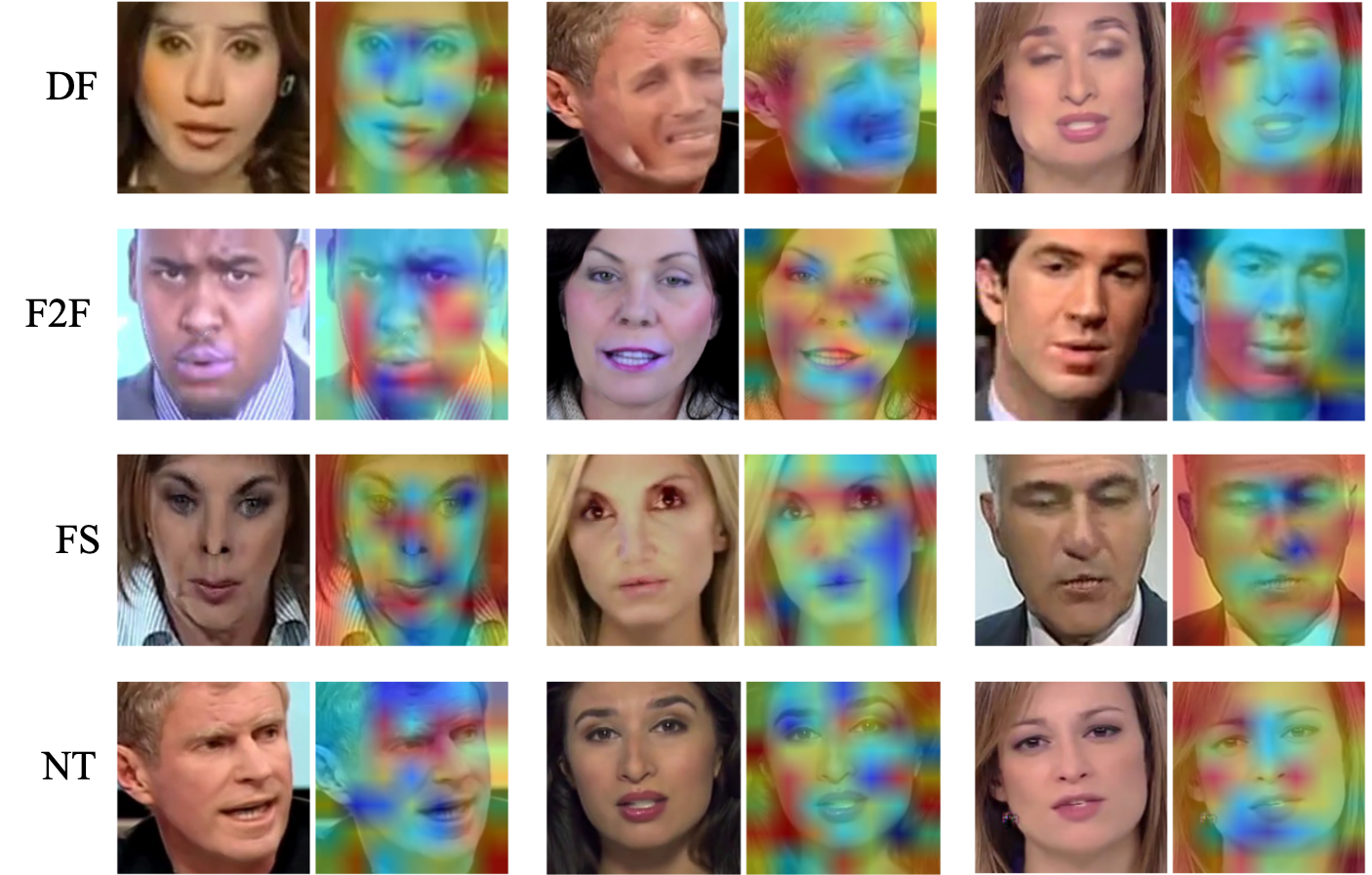}
  \caption{The Grad-CAM~\cite{selvaraju2017grad} Visualization on four manipulations in FF++. The model used to generate Grad-Cam is trained only with DF.}
  \label{fig:gradcam}
\end{figure}

\subsubsection{Generalization to unseen datasets within FF++}
To evaluate the performance of our proposed CFD framework on unseen data, we first conducted a detailed cross-testing analysis by training our model on a particular manipulation technique and evaluating its performance against other techniques listed in FF++. We benchmarked our method against specialized approaches such as MFF~\cite{zhang2024face} and RECCE~\cite{cao2022end}, as presented in \Cref{tab:ff1-3}. 

\begin{table}[htb]
  \caption{Cross-dataset comparison for video-only\textsuperscript{\textasteriskcentered} on FF++(23). The best results are in bold fonts, and the second best is highlighted in blue colour.}
  \label{tab:corssdataset}
  \centering
  \renewcommand{\arraystretch}{1.3} 
  \begin{tabular}{lccccc}
    \hline
    \multirow{2}{*}{Methods} & \multicolumn{3}{c}{Testing Set (AUC)} & \multirow{2}{*}{FLOPs} & \multirow{2}{*}{Params} \\
    \cline{2-4}
     & Celeb-DF & DFDC & DFD \\
    \hline
    LipsForensics~\cite{haliassos2021lips} & 82.4 & {\color{blue} 73.5} & - & - & 24.8M\\
    TALL~\cite{xu2023tall} & \textbf{90.79} & \textbf{76.78} & - & 47.5G & 86M\\
    FTCN~\cite{zheng2021exploring} & 86.9 & 74 & - & - & - \\
    FADE~\cite{tan2023deepfake} & 77.46 & - & \textbf{96.23} & - & - \\
    TI2Net~\cite{liu2023ti2net} & 68.22 & - & 72.03 & - & -\\
    TD-3DCNN~\cite{zhang2021detecting} & 57.32 & 55.02 & - & - & - \\
    VoD (ours) & 72.42 & 65.57 & {\color{blue} 90.99} & 1.96G & 3.76M\\
    \hline
  \end{tabular}
  \parbox{\linewidth}{\footnotesize\textsuperscript{\textasteriskcentered}The SOTA related to methods trained in video-settings are reported and do not include other methods based on image level detection for fair comparison.}
\end{table}

The within-dataset performance of the proposed CFD, indicated by the grey cells, achieves the top performance of each dataset despite a slightly lower score of F2F. This highlights its effectiveness when trained and tested on the same type of manipulation.
Additionally, CFD achieves the highest average AUC when trained on each individual manipulation technique dataset except DF setting, demonstrating a significant advantage. This indicates the robustness and generalization capability of CFD across various types of manipulations.

\begin{figure*}[htb]
      \centering
      \includegraphics[width=0.85\textwidth]{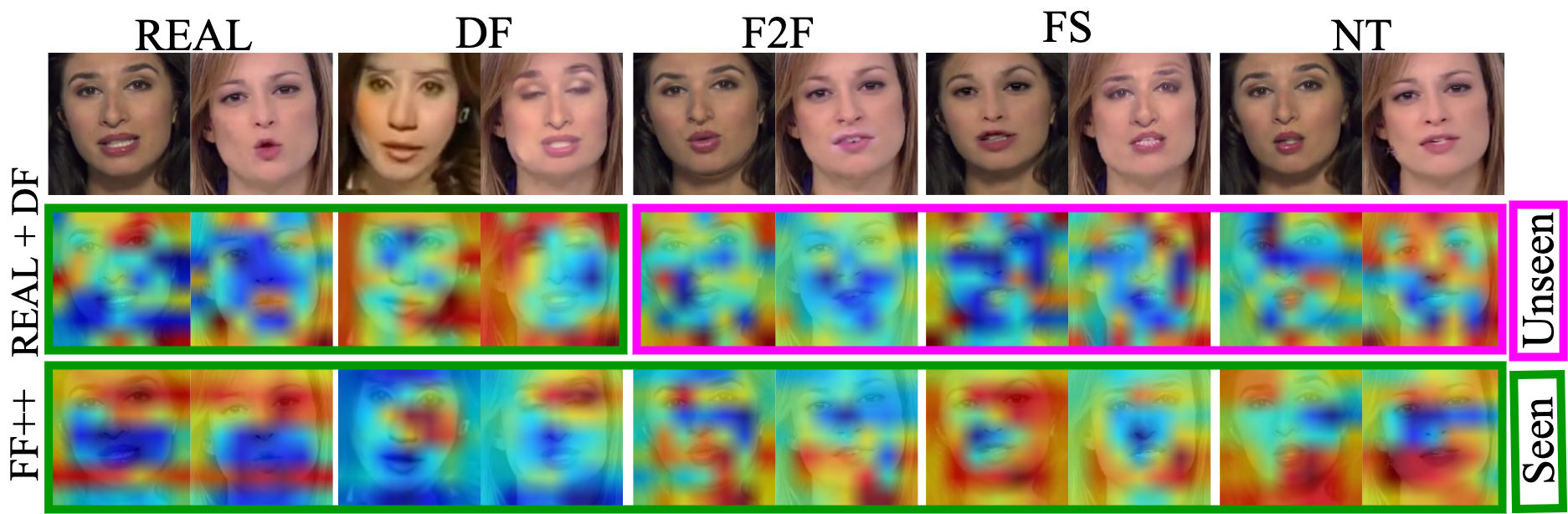}
      \caption{The Grad-CAM~\cite{selvaraju2017grad} Visualization on four manipulations in FF++ for seen and unseen setting.}
      \label{fig:gradcam-seen-unseen}
\end{figure*}

GradCAM visualizations for two models both in seen and unseen data setting is shown in in \Cref{fig:gradcam}. GradCAM highlights reveal that the model primarily focuses on the outlines of the faces when analyzing seen data. When trained on the FF++ dataset, the model exhibits a more nuanced focus with distinct attention regions for each class. The observations suggest that it has learned to recognize a broader array of features across different types of manipulated media.

To better understand the decision-making mechanism of our method on unseen data, we provide the Grad-CAM~\cite{selvaraju2017grad} visualization on FF++ in \Cref{fig:gradcam} where the model is only trained with DF. We observe that for face swap methods (\textit{e.g.}, DF and FS) where the entire face is replaced, our method concentrates on the borders of the faces because segment inconsistencies arise in those areas. While for the face reenactment techniques(\textit{e.g.}, F2F and NT), the focus shifts to the facial region. Remarkably, despite the network being trained solely on DF data, it adeptly generalizes to other manipulation methods. This demonstrates the effectiveness of our VoD method in accurately identifying and focusing on regions of interest specific to different types of facial manipulations.

\begin{table*}[htb]
    \caption{Ablation study of using \textit{video segments} and \textit{CFD} for the training of our CFD framework. AUC score (95\% CI) is reported.}
    \label{tab:abl-video-seg}
    \centering
    \renewcommand{\arraystretch}{1.3}
    \small
    \begin{tabular}{lclcccc}
        \hline
        \multirow{2}{*}{Input} & \multirow{2}{*}{Backbone} & \multicolumn{4}{c}{Train on DF (AUC (95\% CI))} \\ \cline{3-6}
         & & DF & F2F & FS & NT \\
        \hline
        Video seg. & Slow\_r50~\cite{tran2018closer} & 99.37 & 69.02 & 30.71 & 78.43 \\
        \rowcolor{gray!20} & & \footnotesize{(99.06, 99.58)} & \footnotesize{(66.12, 71.71)} & \footnotesize{(28.00, 33.49)} & \footnotesize{(75.89, 80.72)} \\
        CFD & Slow\_r50~\cite{tran2018closer} & 99.66 & 64.02 & \textbf{75.10} $\uparrow$$\uparrow$ & 69.16 \\
        \rowcolor{gray!20} & & \footnotesize{(99.45, 99.79)} & \footnotesize{(61.19, 66.86)} & \footnotesize{(72.46, 77.50)} & \footnotesize{(66.40, 71.82)} \\
        \hline
        Video seg. & X3D-S~\cite{feichtenhofer2020x3d} & 99.70 & 59.02 & 42.18 & 67.74 \\
        \rowcolor{gray!20} & & \footnotesize{(99.45, 99.84)} & \footnotesize{(56.03, 61.89)} & \footnotesize{(39.21, 45.15)} & \footnotesize{(64.96, 70.53)} \\
        CFD & X3D-S~\cite{feichtenhofer2020x3d} & 99.28 & 62.68 & \textbf{78.95} $\uparrow$$\uparrow$ & 66.66 \\
        \rowcolor{gray!20} & & \footnotesize{(98.89, 99.53)} & \footnotesize{(59.72, 65.51)} & \footnotesize{(76.08, 80.79)} & \footnotesize{(63.18, 68.88)} \\
        \hline
    \end{tabular}
\end{table*}

\subsubsection{Cross-dataset detection performance analysis}
Further, to understand the effectiveness of the proposed approach in a cross-dataset setting, another analysis is conducted by training the detection models in FF++(c23) and testing it on unseen data on Celeb-DF, DFDC and DFD. For a fair comparison, we choose the techniques that rely on video-based detection and report the same in \Cref{tab:corssdataset}. When trained on FF++(HQ) and tested on DFD, a clear improvement can be seen ($\approx 91\%$) while the proposed method performs inferior to SOTA for Celeb-DF and DFDC needing further investigations. 
Additionally, the FLOPs and parameters of our method are notably minimal, making our network highly efficient and user-friendly.

\subsection{Robustness}
\begin{figure*}[htb]
  \centering
  \includegraphics[width=0.9\textwidth]{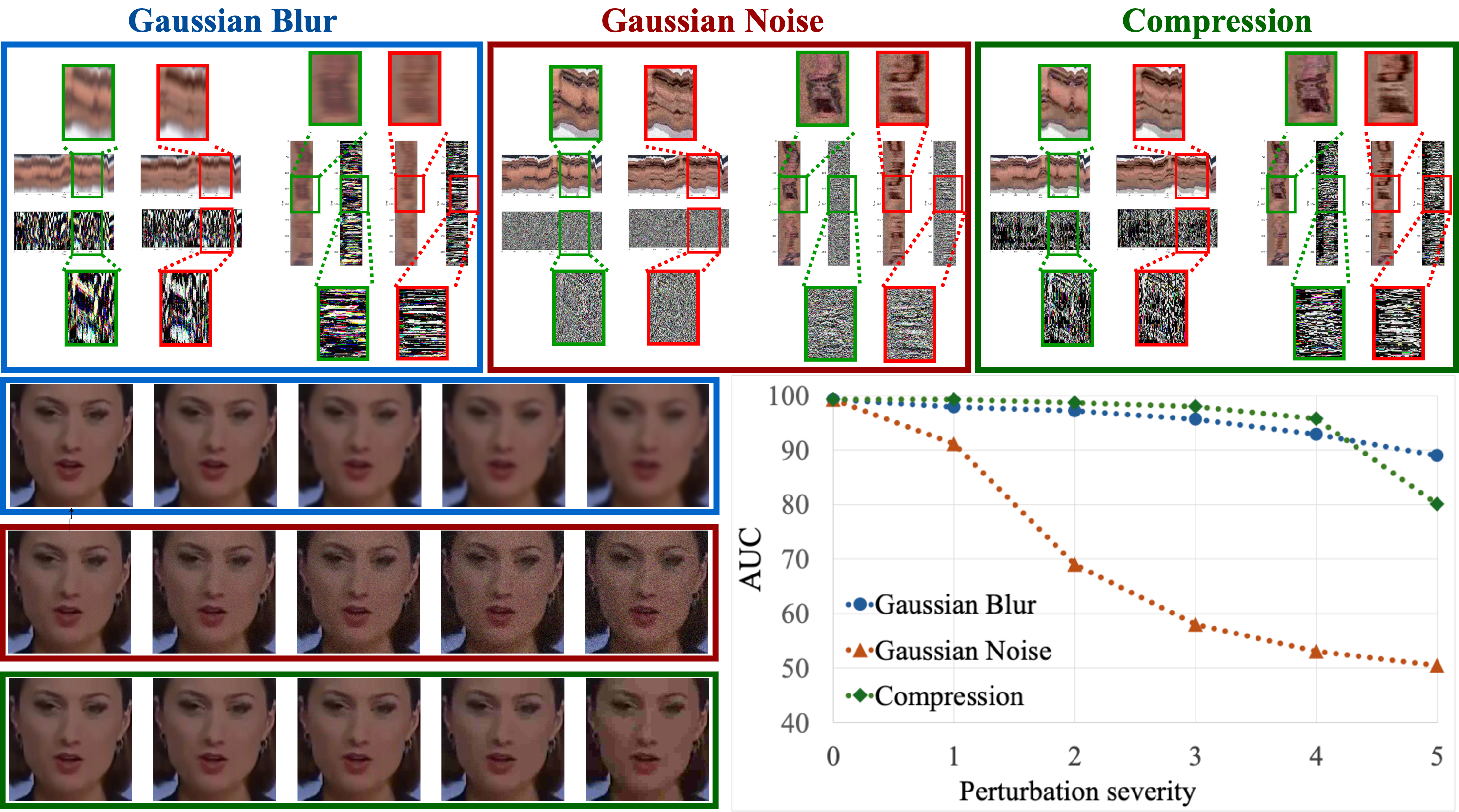}
  \caption{Robustness of proposed approach across different noise factors}
  \label{fig:robustness}
\end{figure*}

Given that simple image noise, unrelated to the videos' authenticity, can affect the calculated difference volumes, we further investigate the robustness of our VoD model against several common types of perturbations. We specifically apply Gaussian blur, Gaussian noise, and compression to the FF++ (HQ) dataset, using five different levels of severity for each type of perturbation. As shown in \Cref{fig:robustness}, our VoD model exhibits significant resilience to Gaussian blur and compression. Even at the highest severity levels, where visual distortion is most severe, the model can identify fake videos with approximately 90\% for Gaussian blur and 80\% for compression regarding the AUC score. However, the model shows greater sensitivity to Gaussian noise, as this noise is introduced randomly on a frame-by-frame basis, significantly impacting VoD performance.

\subsection{Ablation Study}
An ablation study is conducted to evaluate the impact of various components in our consecutive frame difference~(CFD) technique. We compare video segments to the CFD technique and conduct analyses to assess the effects of different segment lengths ($C_{sl}$) and sampling intervals ($C_{in}$), various methods for calculating frame differences, and the use of different backbone networks in this section.

\textbf{Video segment v.s. CFD:}
In order to showcase the effectiveness of the CFD method, we conducted an ablation study using both CFD segments and video segments as inputs, as presented in \Cref{tab:abl-video-seg}. The results indicate that our CFD technique significantly improves network performance compared to using raw video segments, especially on FS, which tends to be the most challenging manipulation technique in FF++~\cite{DBLP:conf/iccv/RosslerCVRTN19, xu2022supervised}. 
Particularly, the performance on the FS dataset shows a significant improvement when using CFD, doubling the score for both slow\_r50 and X3D-S on FS, highlighting the robustness and superiority of our method. These notable performance improvements underscore the effectiveness of our CFD approach in enhancing model accuracy and generalization.

\textbf{Influence of segment length $C_{sl}$:}
Segment length $C_{sl}$ is a crucial parameter in handling temporal information since it sets the scope of temporal analysis. 
\Cref{tab:abl-seg-len} illustrates the effectiveness of various segment lengths in training our CFD method.
The results reveal that a segment length of 24 provides the most balanced performance across different types of manipulations, with the highest AUC score of 78.46. 
Specifically, this segment length achieves an AUC of 83.69 for FaceSwaps, indicating significant adaptability according to the 95\% CI provided in the table.

With the CFD technique, increasing the segment length does not necessarily enhance the detection performance. 
A segment length of 32 yields inferior results compared to lengths of 16 and 24. Additionally, our model is untrainable with a segment length of 64, underscoring the importance of selecting an optimal segment length for efficient model training.
This may occur because the correlation between segments' differences tends to diminish as they become more distant from one another. Therefore, differences with relatively shorter segments could potentially be more informative and useful for training the proposed VoD framework.

\begin{table*}[htb]
    \caption{Ablation study of using different \textit{segment lengths} for the training of our CFD framework. AUC score (95\% CI) is reported.}
    \label{tab:abl-seg-len}
    \centering
    \renewcommand{\arraystretch}{1.3}
    \small
    \begin{tabular}{ccccc>{\centering\arraybackslash}p{2.cm}}
        \hline
        \multirow{2}{*}{Segment length} & \multicolumn{4}{c}{Train on DF (AUC (95\% CI))} & \multirow{2}{*}{Avg} \\ \cline{2-5}
         & DF & F2F & FS & NT & \\
        \hline
        16 & 99.28 & 62.68 & 78.95 & 66.66 & 76.89 \\
        \rowcolor{gray!20} & \footnotesize{(98.89, 99.53)} & \footnotesize{(59.72, 65.51)} & \footnotesize{(76.08, 80.79)} & \footnotesize{(63.18, 68.88)} & \\
        24 & \textbf{99.70} & \textbf{63.16} & \textbf{83.69} & \textbf{67.30} & \textbf{78.46} \\
        \rowcolor{gray!20} & \footnotesize{(99.53, 99.82)} & \footnotesize{(60.31, 66.07)} & \footnotesize{(81.59, 85.61)} & \footnotesize{(64.46, 70.03)} & \\
        32 & 99.11 & 59.49 & 80.76 & 59.93 & 74.82 \\
        \rowcolor{gray!20} & \footnotesize{(98.62, 99.43)} & \footnotesize{(56.43, 62.42)} & \footnotesize{(78.41, 82.89)} & \footnotesize{(56.94, 62.95)} & \\
        \hline
    \end{tabular}
\end{table*}

\textbf{Influence of sampling interval $C_{in}$: }
Sampling interval $C_{in}$ is another critical factor in the training of the VoD framework.
It represents the number of frames skipped between each selected frame during the training of the backbone model, providing insight into the most informative frequency and interaction between frames.
The performance of difference $C_{in}$ is showing in \Cref{tab:abl-sam-in}.
The results indicate that a sampling interval of 1 yields the highest performance for the CFD technique, showcasing that the consecutive frame difference contains the most critical information for Deepfake detection.
As the sampling interval increases, there is a noticeable decline in performance. 
This trend suggests that larger intervals omit critical temporal information, which reduces the model's generalizability. 
As a result, the model becomes less capable of accurately detecting Deepfake across varied manipulation techniques due to the loss of essential frame-to-frame details.

\begin{table*}[htb]
    \caption{Ablation study of using different \textit{sampling intervals} for the training of our CFD framework. AUC score (95\% CI) is reported.}
    \label{tab:abl-sam-in}
    \centering
    \renewcommand{\arraystretch}{1.3}
    \small
    \begin{tabular}{ccccc>{\centering\arraybackslash}p{1.5cm}} 
        \hline
         & \multicolumn{4}{c}{AUC with train dataset as DF} & \multirow{2}{*}{Avg} \\ 
        \multirow{2}{*}{Frame rate} & \multicolumn{4}{c}{(95\% CI indicated in parenthesis)} & \\
        \cline{2-5}
         & DF & F2F & FS & NT & \\
        \hline
        1 & \textbf{99.28} & \textbf{62.68} & \textbf{78.95} & \textbf{66.66} & \textbf{76.89} \\
        \rowcolor{gray!20} & \footnotesize{(98.89, 99.53)} & \footnotesize{(59.72, 65.51)} & \footnotesize{(76.08, 80.79)} & \footnotesize{(63.18, 68.88)} & \\
        2 & 98.96 & 57.78 & 71.36 & 57.28 & 71.35 \\
        \rowcolor{gray!20} & \footnotesize{(98.30, 99.37)} & \footnotesize{(54.45, 61.13)} & \footnotesize{(68.34, 74.36)} & \footnotesize{(53.91, 60.54)} & \\
        3 & 99.15 & 56.96 & 67.93 & 57.08 & 70.28 \\
        \rowcolor{gray!20} & \footnotesize{(98.67, 99.45)} & \footnotesize{(53.50, 60.22)} & \footnotesize{(64.87, 70.95)} & \footnotesize{(53.68, 60.36)} & \\
        4 & 99.12 & 59.79 & 66.10 & 58.92 & 70.98 \\
        \rowcolor{gray!20} & \footnotesize{(98.69, 99.42)} & \footnotesize{(56.48, 63.09)} & \footnotesize{(62.93, 69.20)} & \footnotesize{(55.60, 62.20)} & \\
        5 & 98.25 & 57.74 & 69.14 & 56.51 & 70.41 \\
        \rowcolor{gray!20} & \footnotesize{(97.51, 98.76)} & \footnotesize{(54.40, 61.14)} & \footnotesize{(66.03, 72.18)} & \footnotesize{(53.19, 59.76)} & \\
        6 & 98.88 & 62.39 & 64.36 & 62.59 & 72.06 \\
        \rowcolor{gray!20} & \footnotesize{(98.38, 99.23)} & \footnotesize{(59.03, 65.52)} & \footnotesize{(61.05, 67.53)} & \footnotesize{(59.36, 65.87)} & \\
        \hline
    \end{tabular}
\end{table*}

\textbf{Successive Subtraction by First Frame v.s. CFD}
We attempt to use the Successive Subtraction by First Frame (SSFF) strategy, where the first frame is fixed and substracted. 
Unlike the CFD method, which calculates the difference between consecutive frames, the SSFF approach calculates the difference between each frame and the first frame of the segment, expressed as $|F_x - F_1|, x \in [2, C_{sl}]$.
The outcomes depicted in \Cref{tab:abl-ssff} illustrate that the CFD segments notably outperform SSFF segment across all the manipulations, reinforcely suggesting that consecutive frame differences with $C_{in}$ equals to $1$ have more valuable temporal information for accurate Deepfake detection. 

\begin{table*}[htb]
    \caption{Ablation study of using \textit{successive subtraction} and \textit{CFD} for the training. AUC score (95\% CI) is reported.}
    \label{tab:abl-ssff}
    \centering
    \renewcommand{\arraystretch}{1.3}
    \small
    \begin{tabular}{ccccc>{\centering\arraybackslash}p{1.5cm}}
        \hline
        \multirow{2}{*}{\shortstack{Subtraction \\Pattern}} & \multicolumn{4}{c}{Train on DF (AUC (95\% CI))} & \multirow{2}{*}{Avg} \\ \cline{2-5}
         & DF & F2F & FS & NT & \\
        \hline
        SSFF & 96.11 & 55.43 & 69.49 & 61.26 & 69.00 \\
        \rowcolor{gray!20} & \footnotesize{(95.02, 97.01)} & \footnotesize{(52.41, 58.45)} & \footnotesize{(66.61, 72.10)} & \footnotesize{(58.33, 64.13)} & \\
        CFD~(Ours) & \textbf{99.28} & \textbf{62.68} & \textbf{78.95} & \textbf{66.66} & \textbf{76.89} \\
        \rowcolor{gray!20} & \footnotesize{(98.89, 99.53)} & \footnotesize{(59.72, 65.51)} & \footnotesize{(76.08, 80.79)} & \footnotesize{(63.18, 68.88)} & \\
        \hline
    \end{tabular}
\end{table*}

\textbf{Various Backbone Networks.}
In this part, we perform an ablation study to assess the impact of various backbone networks on the performance of the proposed VoD framework. 
The study includes six backbone networks: Slow\_r50~\cite{tran2018closer}, EfficientNet\_X3d~\cite{feichtenhofer2020x3d}, MViT(16×4)~\cite{fan2021multiscale}, Slowfast\_r50 and Slowfast\_r50~(+frame)\cite{feichtenhofer2019slowfast}, and X3D-S~\cite{feichtenhofer2020x3d}. 
As the Slowfast network requires two-stream inputs, we test two different input combinations: one using slow segments of CFD and fast segments of CFD, and the other combining slow segments from original frames and fast segments of CFD.

Comparing the performance across different manipulations (DF, F2F, FS, NT) in FF++ as shown in \Cref{tab:abl-backbones}, the X3D-S backbone outperforms others, achieving an average AUC score of 75.34, indicating its superior performance. Notably, it performs exceptionally well on the DF (99.67\%) and FS (80.89\%) datasets. The Slow\_r50 backbone shows strong results on the DF dataset with an AUC of 99.15\% and on the NT dataset with an AUC of 70.44\%. 

Our proposed VoD framework performs better with simpler backbone networks like X3D-S and slow\_r50. However, when used with more complex networks like MViT and Slowfast, the performance does not show significant improvement and, in some cases, even decreases. This suggests that our VoD framework may generate more generalizable features with simpler networks.

\begin{table}[htb]
    \caption{Ablation study of using different \textit{backbone networks} for the training of VoD. SGD optimizer is used for all the training. AUC score is reported.}
    \label{tab:abl-backbones}
    \centering
    \renewcommand{\arraystretch}{1.3} 
    \resizebox{0.48\textwidth}{!}{
    \normalsize
    \begin{tabular}{lcccc>{\centering\arraybackslash}p{1.3cm}} 
        \hline
        \multirow{2}{*}{Backbone} & \multicolumn{4}{c}{Train on DF} & \multirow{2}{*}{Avg} \\ \cline{2-5}
         & DF & F2F & FS & NT & \\
        \hline
        Slow\_r50~\cite{tran2018closer} & 99.15 & \textbf{64.37} & 61.6 & \textbf{70.44} & 73.89 \\
        EfficientNet\_X3d~\cite{feichtenhofer2020x3d} & 99.52 & 58.83 & 76.97 & 63.01 & 74.58 \\
        MViT($16\times4$)~\cite{fan2021multiscale} & 98.86 & 57.79 & 77.77 & 59.66 & 73.31 \\
        Slowfast\_r50~\cite{feichtenhofer2019slowfast} & 94.60 & 53.95 & 60.35 & 51.64 & 65.14 \\
        Slowfast\_r50(+frame)~\cite{feichtenhofer2019slowfast} & 99.71 & 52.17 & 67.45 & 47.73 & 66.77 \\
        X3D-S~\cite{feichtenhofer2020x3d} & \textbf{99.67} & 60.02 & \textbf{80.89} & 60.78 & \textbf{75.34} \\
    \hline
    \end{tabular}
    }
\end{table}

\section{Conclusion}
Volume of Differences (VoD) introduced in this work detects Deepfakes  by utilizing temporal and spatial inconsistencies between consecutive video frames. Specifically, we utilized the Consecutive Frame Difference~(CFD) to obtain the input volume of differences and expendable networks for classification.
The temporal discrepancies between consecutive frames helps specifically in discovering subtleties for classifying real and forged videos.
The proposed approach is evaluated on several well-known publicly available Deepfake datasets in intra-dataset and cross-dataset testing scenarios.
The results demonstrate that the VoD framework not only excels in the seen data but also has robust generalizability on unseen data.
Additionally, we conducted a thorough ablation study to determine the optimal performance parameters within our VoD framework, further enhancing its efficacy and applicability.





%


\bibliographystyle{ieee}
\bibliography{main}

\begin{IEEEbiography}[{\includegraphics[width=1in,height=1.25in,clip,keepaspectratio]{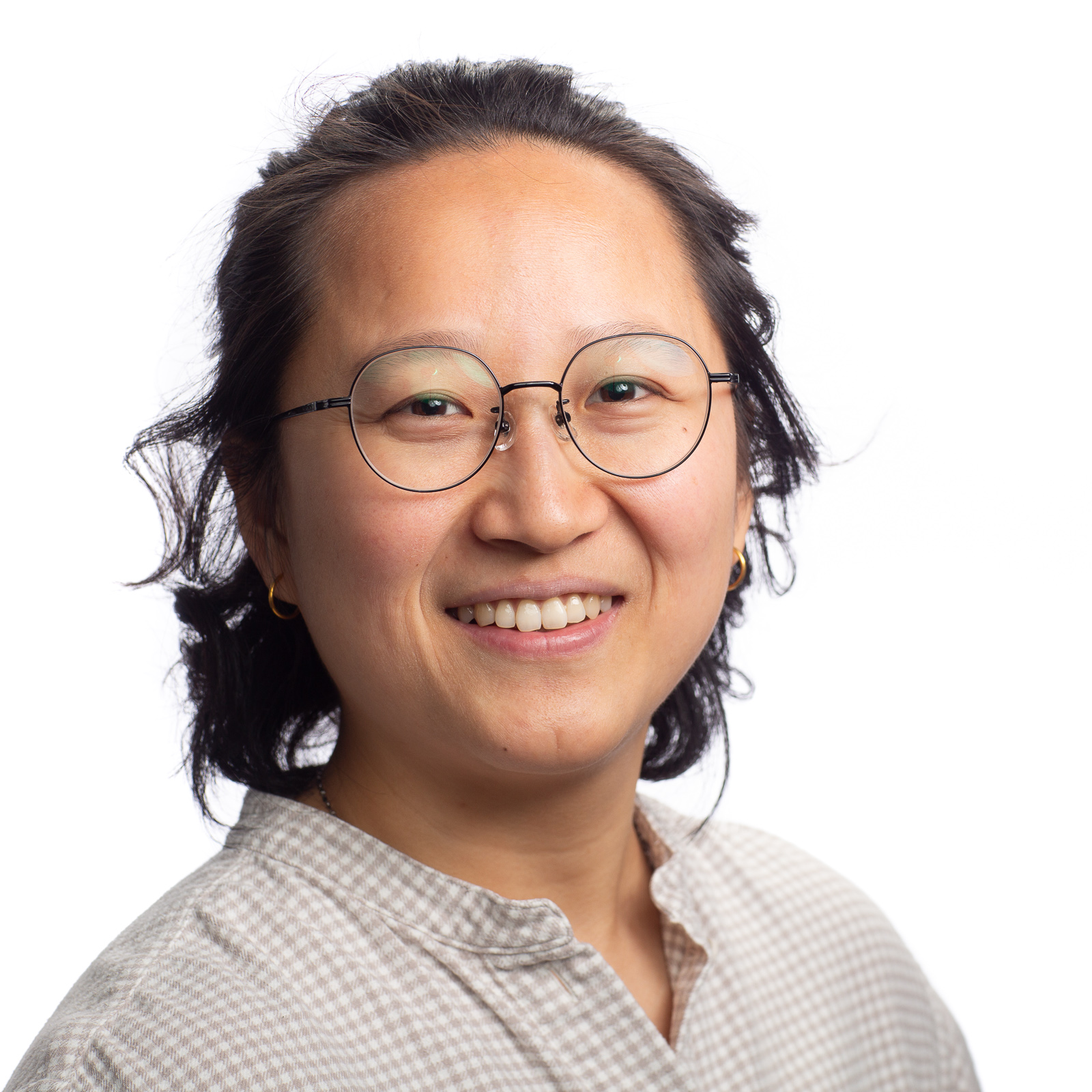}}]{Ying Xu}
received her B.Sc. in Electrical Engineering in 2015 from Shanghai University, China. She completed her M.Sc. in Applied Computer Science in 2021 from Norwegian University of Science and Technology, Norway, where she is currently pursuing the Ph.D. degree, focusing on Deepfake detection.
\end{IEEEbiography}

\vspace{-11pt}

\begin{IEEEbiography}[{\includegraphics[width=1in,height=1.25in,clip,keepaspectratio]{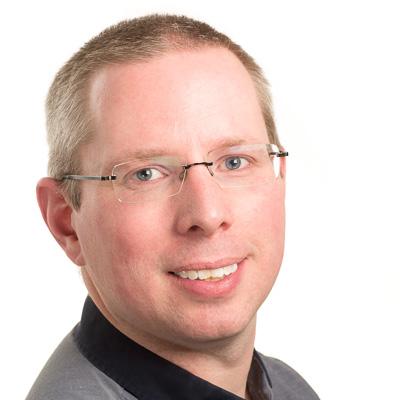}}]{Marius Pedersen}
received his BsC in Computer Engineering in 2006, and MiT in Media Technology in 2007, both from Gjøvik University College, Norway. He completed a PhD program in color imaging in 2011 from the University of Oslo, Norway, sponsored by Océ. He is professor at the Department of Computer Science at NTNU in Gjøvik, Norway. He is also the director of the Norwegian Colour and Visual Computing Laboratory (Colourlab). His work is centered on subjective and objective image quality.
\end{IEEEbiography}

\vspace{-11pt}

\begin{IEEEbiography}[{\includegraphics[width=1in,height=1.25in,clip,keepaspectratio]{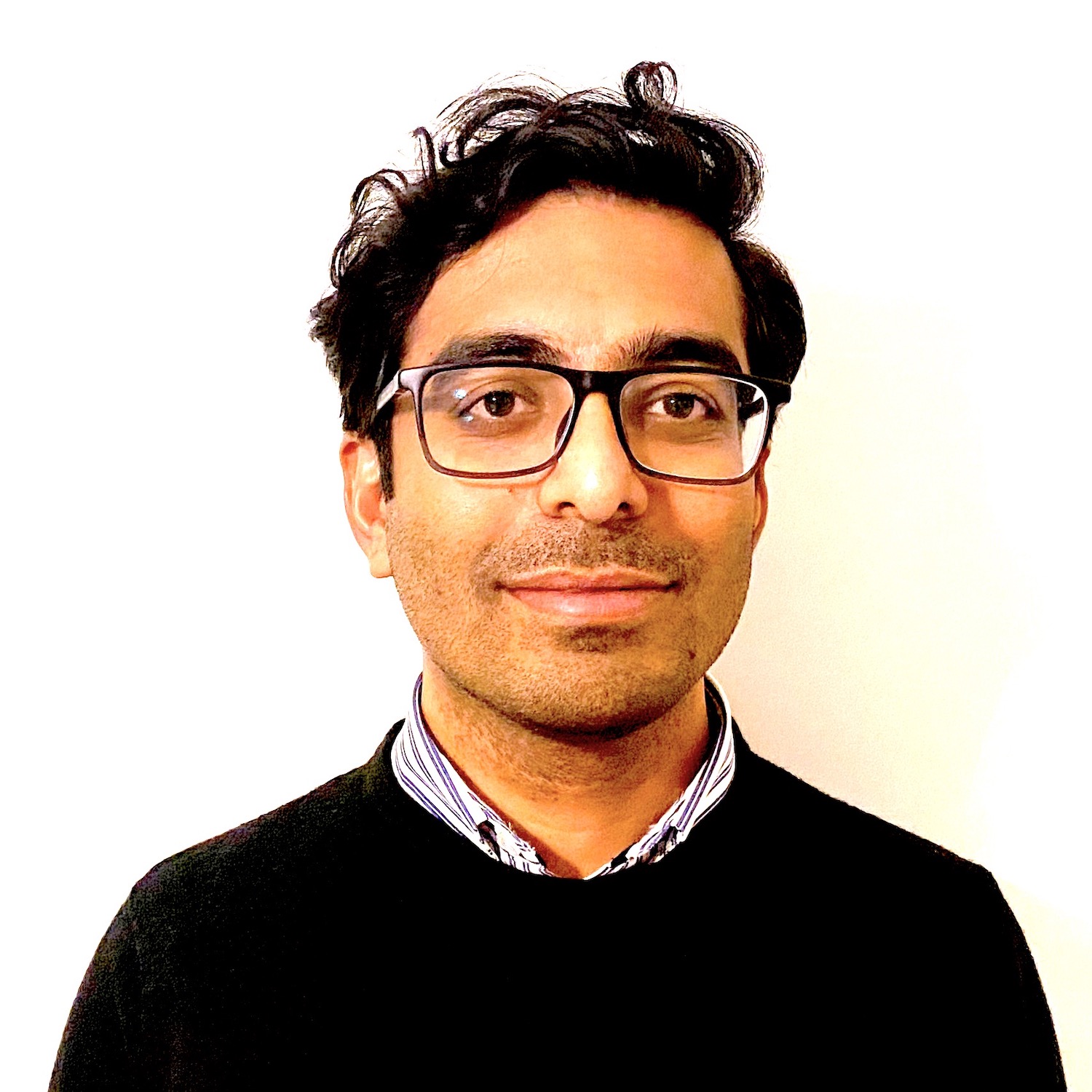}}]{Kiran Raja} (Senior Member, IEEE) received the Ph.D. degree in computer Science from the Norwegian University of Science and Technology, Norway, in 2016, where he is Faculty Member with the Department of Computer Science. His main research interests include statistical pattern recognition, image processing, and machine learning with applications to biometrics, security and privacy protection. He was/is participating in EU projects SOTAMD, iMARS, and other national projects. He is a member of European Association of Biometrics (EAB) and chairs Academic Special Interest Group at EAB. He serves as a reviewer for number of journals and conferences. He is also a member of the editorial board for various journals.
\end{IEEEbiography}

\end{document}